Research paper

# Your turn: At home turning angle estimation for Parkinson's disease severity assessment

Qiushuo Cheng [a,*], Catherine Morgan [b,c], Arindam Sikdar [a], Alessandro Masullo [a], Alan Whone [b,c], Majid Mirmehdi [a]

[a] Faculty of Engineering, University of Bristol, UK
[b] Translational Health Sciences, University of Bristol, UK
[c] North Bristol NHS Trust, Southmead Hospital, Bristol, UK

ABSTRACT

People with Parkinson's Disease (PD) often experience progressively worsening gait, including changes in how they turn around, as the disease progresses. Existing clinical rating tools are not capable of capturing hour-by-hour variations of PD symptoms, as they are confined to brief assessments within clinic settings, leaving gait performance outside these controlled environments unaccounted for. Measuring turning angles continuously and passively is a component step towards using gait characteristics as sensitive indicators of disease progression in PD. This paper presents a deep learning-based approach to automatically quantify turning angles by extracting 3D skeletons from videos and calculating the rotation of hip and knee joints. We utilise advanced human pose estimation models, Fastpose and Strided Transformer, on a total of 1386 turning video clips from 24 subjects (12 people with PD and 12 healthy control volunteers), trimmed from a PD dataset of unscripted free-living videos in a home-like setting (Turn-REMAP). We also curate a turning video dataset, Turn-H3.6M, from the public Human3.6M human pose benchmark with 3D groundtruth, to further validate our method. Previous gait research has primarily taken place in clinics or laboratories evaluating scripted gait outcomes, but this work focuses on free-living home settings where complexities exist, such as baggy clothing and poor lighting. Due to difficulties in obtaining accurate groundtruth data in a free-living setting, we quantise the angle into the nearest bin 45° based on the manual labelling of expert clinicians. Our method achieves a turning calculation accuracy of 41.6%, a Mean Absolute Error (MAE) of 34.7°, and a weighted precision (WPrec) of 68.3% for Turn-REMAP. On Turn-H3.6M, it achieves an accuracy of 73.5%, an MAE of 18.5°, and a WPrec of 86.2%. This is the first work to explore the use of single monocular camera data to quantify turns by PD patients in a home setting. All data and models are publicly available, providing a baseline for turning parameter measurement to promote future PD gait research.

## 1. Introduction

Parkinson's disease (PD) is a progressive neurodegenerative movement disorder, characterised by symptoms such as slowness of movement and gait dysfunction [1] which fluctuate across the day but progress slowly over the years [2]. Currently, treatment of PD relies on therapies which improve symptoms. There are no treatments available which modify the course of the underlying disease (so-called disease-modifying treatments, or DMTs), despite there being multiple putative DMTs showing promise in laboratory studies [3]. One reason for the slow development of DMTs is the dearth of sensitive, frequent, objective biomarkers to enhance the current gold-standard clinical rating scale [4] to measure the progression of PD. This gold-standard clinical rating scale, the Movement Disorders Society-sponsored revision of the Unified Parkinson's Disease Rating Scale (MDS-UPDRS) [4], includes subjective questionnaires concerning gait and mobility experiences, along with clinicians' ratings of scripted activities performed by the participants. The assessments typically occur within clinical settings over short durations, offering only a "snapshot" of symptoms which vary on an hourly basis. It also has limitations including its non-linear and discontinuous scoring system, the inter-rater variability [5] and Hawthorne effect [6] of being observed on how someone mobilises [7, 8].

Gait and turning abnormalities are common features of PD with over half of the patients reporting difficulties with turning [9] —

* Corresponding author.
*E-mail address:* wl22741@bristol.ac.uk (Q. Cheng).

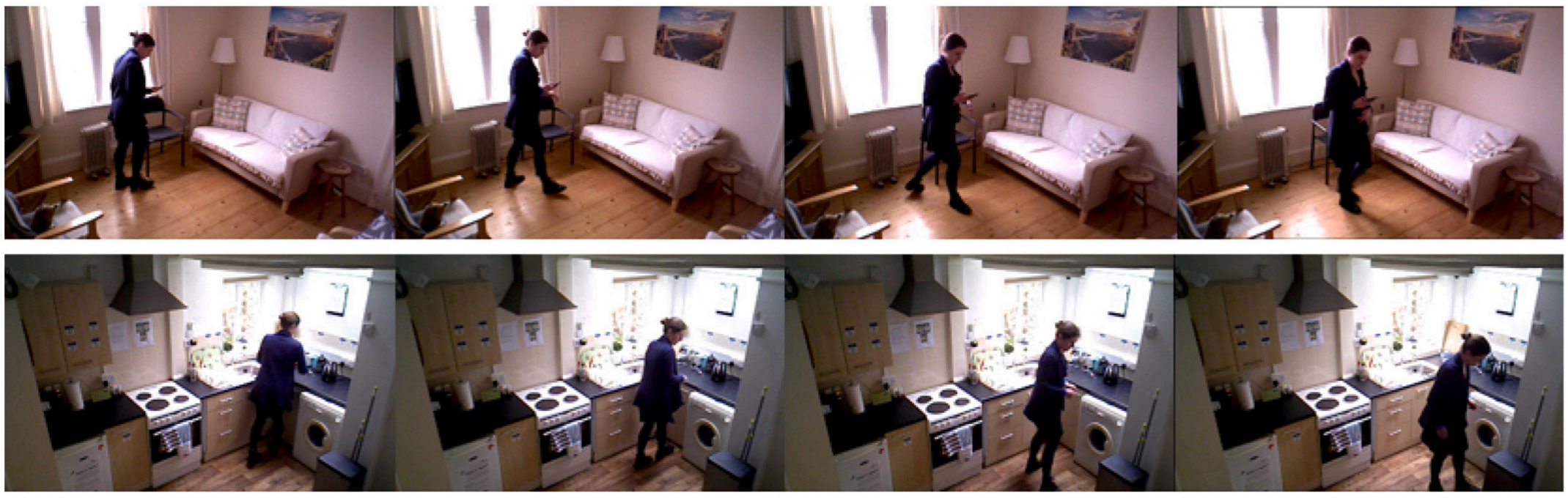

**Fig. 1.** Examples of free-living home scenario turning activity in REMAP [18] — Sample frames (first row) in the living room and (second row) in the kitchen.

when someone moves round on their axis while upright, changing the direction they face. Turning changes associated with PD include the 'en bloc' phenomenon where upper and lower body segments turn simultaneously [10], a longer duration of turn, less accurate turn completion, a narrower base of support [11] and the use of 'step turns' rather than 'pivot turns' [12]. More than 40% of daily steps are during turning [13] and turning abnormalities can predispose to falls, thus turning parameters could potentially be used as measures predicting the time to falls in a patient with PD [9]. Furthermore, if a fall happens during turning, it is up to 8 times more likely to result in a hip fracture [14]. In unmedicated early-stage PD, gait parameters from turning are more sensitive to change compared to straight-ahead gait outcomes [15], making measuring aspects of turns potentially of specific use in clinical trials of disease-modifying interventions which typically recruit recently diagnosed patients [16]. People with PD turn differently when being watched by a clinician [17], so measuring turning passively in uncontrolled home settings (Fig. 1) could give information about mobility not captured by face-to-face assessments in the clinic.

Measuring turning angle has clinical value in Parkinson's for several reasons. Firstly, it could detect whether someone has PD (compared to a healthy person) from in-home data, since people with PD take turns with smaller angles and show more variability in their turning angles throughout the day and week than control subjects [19]. Secondly, it can be used to better understand the symptom severity in PD. Turning angles have been shown to illustrate PD medication status, since people with PD take larger angles of turn when they are taking medications compared to when they withhold them [20]. Turning angles could also be used to detect whether people with PD have the symptom of "freezing of gait", since those subjects take on average smaller angles of turn and fewer large amplitude/angle turns at home than control subjects [21]. Thirdly, detected turning angles can be used to predict fall risk (tested in healthy subjects), since the mean frequency of falls has been shown to be correlated with the angle of turn [22]. Fourthly, knowing what angle a turn has been taken over can be vital in detecting and understanding any other measurements taken from gait/turning — for instance, comparing like for like, 180° turns can reveal early signs of disease progression in PD [23]. Being able to detect and isolate larger angle turns would improve precision in demonstrating specific PD gait features, for example, subjects who freeze in gait show increased step time variability and increased numbers of freezing episodes at larger turning angles [24]. Finally, using simple logic, the knowledge of turning angle is vital to contextualise other turning parameters such as duration, step time variability and speed of turn in Parkinson's disease — for example, knowing whether a turn was over 90° or 360° would add vital clinical information to the quantification of that turn's duration.

In this paper, we present a deep learning-based pipeline to estimate turning angles. We adopt advanced human pose estimation models to extract 3D human body joint coordinates from monocular RGB videos. The angle of the turn can be calculated by leveraging the orientation of the paired (left and right) hip and knee joints, which are on the frontal plane of the human body and serve as reliable indicators of the direction in which the body is facing. We apply the proposed pipeline on Turn-REMAP, a dataset of turning video clips trimmed from the unique REMAP dataset [18,25], which includes unscripted spontaneous turning activities from passively collected home monitoring videos. To evaluate our proposed method, a retrospective analysis of the trimmed video clips by clinicians serves as the groundtruth reference. As it is hard to acquire the precise degree of turning using the naked eye, we adopted a special quantisation method: different from the reference technique used by previous studies [26,27], we classify turning angles into the nearest discrete 45° bins.

To the best of our knowledge, this study is the first to use computer vision technology to measure turning angles in free-living videos for people with PD without relying on the traditional gold standard of motion capture reference typically used in laboratory settings. We also curate Turn-H3.6M comprising 619 turning clips trimmed from the public benchmark Human3.6M [28], obtained under controlled settings, and we apply the same turning angle calculation pipeline for comparative evaluation. Due to the availability of 3D data in Turn-H3.6M, we can also compute the turning speed.

In summary our contributions are as follows:

- We introduce the Turn-REMAP dataset which provides the first collection of free-living turning videos recorded in a home environment, with both PD patients and healthy controls. The dataset includes discretised groundtruth turning angles generated by expert clinicians.
- We curate turning videos from the large-scale laboratory-based Human3.6M [28] benchmark dataset which includes motion-capture groundtruth.
- Utilising human pose estimation models, we propose a pipeline to estimate turning angles from single-view RGB videos and validate this pipeline on our proposed datasets. This is the first work to estimate turning angles from natural free-living video data captured from people living with PD.

Next, in Section 2, we review the literature which identifies the gap in current PD gait research and provides the context for our contributions in using free-living video-based settings. Section 3 provides a detailed introduction to our datasets, Turn-REMAP and Turn-H3.6M. Then, Section 4 introduces our methodology for the turning angle estimation pipeline, followed by Section 5 which provides implementation details and the evaluation results of the proposed method. Section 6 includes ablation studies to examine the effect of different design choices within the pipeline. In Section 7, we provide a detailed discussion of the experiment results and highlight the novelty and the contribution of our work. Finally, we present our conclusion and outline potential future work that can be built upon these datasets and the proposed baseline methods in Section 8. Our turning measuring algorithms and extracted skeletons on our datasets are available at *link will be provided if published*.

## 2. Literature review

In this section, we consider related literature in the two most pertinent aspects of our work, i.e. sensor-based turning angle estimation and human pose estimation in gait analysis

**Turning angle estimation —** To acquire objective quantitative turning parameters for human motion analysis, inertial sensors which consist of gyroscopes and accelerometers [19,21,26,29–34] and floor pressure sensors [27,35–37] have been well-explored over the years. Many algorithms using inertial sensors placed on shoes or belts have been validated with gold-standard motion capture systems or human raters with reported accuracy on a sub-degree level [19,26,29] but limited to a laboratory environment on the scripted turning course with few predefined turns. Additionally, even though sensors give nearly groundtruth readings under these restricted conditions, they require digital devices to be worn on the body of the participant which raises issues of acceptability [38] and usability [39]. The portable wearable sensors for gait evaluation are power-thirsty and have limited memory storage space, and therefore there are significant burdens for both participants and professionals to replace, recharge, re-configure and transfer data manually. This hinders the generalisation ability of their proposed methods to different patient cohorts, especially in the free-living environment where it is hard to control every relevant factor, like the imperfect use and configuration of wearables. It has been shown that sensor-based algorithms evaluating gait translate poorly from laboratory to home [40]. Furthermore, several papers have demonstrated that people mobilise differently in the laboratory compared to home settings [7,8,31,32].

Another inherent limitation of wearable-based methods is that they can only provide kinematics parameters on a few body parts, rather than a holistic view of the position and orientation of the entire body. Furthermore, it is shown in [34] that placing wearables on different locations of the body (head, neck, lower back and ankle) causes inconsistency in estimated turning angles. Video-based markerless approaches [27,41,42] present a passive and less obtrusive solution to these innate problems of wearable-based approaches. However, compared to marker-based approaches, the accuracy [43] of joint angle estimation is not yet good enough for clinical application. The work in [43,44] has shown that the reported performances are inconsistent and hard to reproduce outside laboratory environments and among different patient cohorts, as they often use off-the-shelf human pose analysis software and hardware on experiments set up in restricted laboratories with scripted activities. To develop and validate robust video-based gait analysis algorithms, the challenge lies in acquiring videos that are representative enough across different patients in different scenarios including clinics, hospitals and homes. We gather and annotate a dedicated, free-living video data set to estimate turning angles to complement existing research on gait analysis for PD.

**Gait Analysis for Parkinson's Disease —** Gait analysis plays an instrumental role in many clinical applications, and is studied closely in PD [45,46]. With widely available open-source pose estimation models applied to movement videos collected during clinical assessments, most leading approaches in the field analyse such patient videos using deep learning models and compare their outcomes against clinicians' annotations to establish the clinical meaning of the measured gait features.

Sato et al. [47] used Openpose [48] to extract skeleton keypoints and designed handcrafted features to measure step cadence. They do not perform any quantitative evaluations on the accuracy of their measured parameters and only provide correlation analysis of their measured gait features with MDS-UPDRS gait scores. Rupprechter et al. [49] also applied Openpose [48] and hand-crafted features from extracted skeletons, but on a large-scale video dataset of hundreds of PD patients, topped with a machine learning classifier to output MDS-UPDRS scores. Similarly, Sabo et al. [50] used Kinect-generated 3D data with clinical annotations to fine-tune ST-GCN [51] to regress the MDS-UPDRS gait scores, whereas the model was originally designed for the task of action recognition. Lu et al. [52] developed and trained their own deep learning model using self-recorded gait examination videos along with similar gait videos from the CASIA Gait Database [53] to extract 3D skeletons and predict MDS-UPDRS gait scores in an end-to-end manner. Guo et al. [54] developed a graph convolutional neural network to predict gait scores of MDS-UPDRS for 157 PD participants. Mehta et al. [55] deployed existing pose estimation models [56,57] to extract 3D skeletons from sit-to-stand videos, and then tested several deep learning models [51,58,59] to infer MDS-UPDRS sub-scores of gait disorder and bradykinesia.

Inferring subscores of MDS-UPDRS from videos potentially could contribute to early diagnosis, remote screening of PD and enable more frequent self-assessment. However, the current studies are still confined within the limitations of traditional clinical rating scales. As shown in [60], the changes in MDS-UPDRS at baseline, month 6, and month 12 could not reflect the decline in gait speed caused by PD, compared to healthy control. This effectively proves that the traditional rating scale cannot sensitively detect the progression of disease which evolves slowly over the years. Therefore, to complement the current clinical tools, it is necessary to quantitatively validate the accuracy of measured gait parameters, and then translate the output to the associated clinical annotations.

We argue that for PD, it is important to accurately measure absolute gait parameters like turning angle, in a natural, non-hospital setting, to build sufficiently sensitive markers to reliably track changes in gait throughout the day and over the years. To this end, most of the previous pose-based research on gait analysis with angular measurement focuses on sagittal [41,61] or coronal [27,62] joint angles. Only [27,63,64] use pose estimation algorithms on turning analysis, but all three are solely for turn detection, or other gait metrics like step length rather than turning angle. Also, other sensors like 3D depth cameras [65] have been used for human pose estimation using depth maps or point clouds. While such an approach is optimal in some applications like virtual reality, there are many unresolved challenges such as being unable to handle self-occlusions or multi-person detection [66]. In the complex, unscripted scenario of monitoring everyday activities, it may not be suitable to use depth sensors alone; however, combining it with RGB data could potentially lead to better results.

## 3. Datasets

Our proposed turning angle measurement approach is evaluated on the turning scenes of the recently released free-living dataset, REMAP [18], and a curated dataset extracted from the public pose estimation benchmark Human3.6M [28]. In this section, we discuss the details of the video data and how our annotations enable quantitative evaluation of our method.

**Turn-REMAP —** REMAP [18] includes PD and healthy participants engaging in actions, such as sit-to-stand transitions or walking turns within a home environment. These specific actions were recorded during undirected and free-living situations, as well as formal clinical evaluations. We present Turn-REMAP, a subset of this data comprising all its turning actions, loosely-scripted and spontaneous (see Fig. 1). The video data is collected using Microsoft Kinect wall-mounted cameras installed on the ground floor (communal areas) of a test-bed house [67] which captured red-green-blue (RGB) and depth data 2–3 h daily (during daylight hours at times when participants were at home). The acceptability of using such high-resolution video recordings for validation purposes in home settings in PD has been studied in [68,69]. Table 1 summarises the details of Turn-REMAP. The dataset contains 12 spousal/parent–child/friend-friend pairs (24 participants in total) living freely in this sensor-embedded smart home for five days at a time. Each pair consists of one person with PD and one person who was a healthy control volunteer (C). This pairing was chosen to enable PD vs.

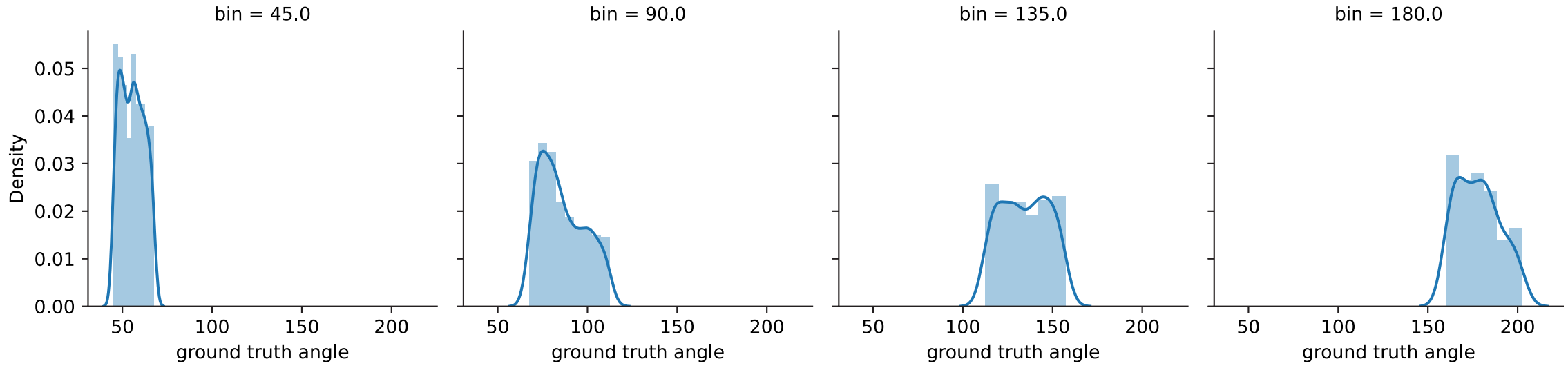

**Fig. 2.** Quantised distribution of angles — Angles are quantised into 45° bins, e.g. a turning angle anywhere in the range 90° ± 22.5° is labelled as 90°. The distribution plot on the graph is from Turn-H3.6M. The groundtruth turning angle on the $x$-axis in this graph is calculated by the changes in body orientation, represented by two hip joints projected onto the ground plane in the 3D motion capture data.

**Table 1**
Summary of Turn-REMAP details. C stands for control subjects.

| #Videos | #Frames | #PD | #C | Avg. PD Age | Avg. C Age | Avg. Age | Avg. Time Since Diag. |
|---|---|---|---|---|---|---|---|
| 1386 | 96 984 | 12 | 12 | 61.25 | 59.25 | 60.25 | 11.3 years |

C comparison, for safety reasons and also to increase the naturalistic social behaviour (particularly amongst the spousal pairs who already lived together). Of the 24 participants, five females and seven males have PD. The average age of the participants is 60.25 (PD 61.25, Control 59.25) and the average time since PD diagnosis for the person with PD is 11.3 years (range 0.5–19). The PD participants were asked to withhold their dopaminergic medications (12 h for short-acting levodopa-containing agents and 24 h for long-acting therapies) and/or switch off their deep brain stimulators (DBS) so that they were in the practically-defined "off" medications/DBS state for a limited period of hours [18]. As the PD participants are in mild to moderate stages of the disease, freezing-of-gait is less evident in our dataset.

The RGB videos were watched post-hoc by medical doctors who had undertaken training in the MDS-UPDRS rating score, including gait parameter evaluation. Two clinicians watched up to 4 simultaneously captured video files at a time using ELAN software [70] to manually annotate the videos to the nearest millisecond to the extent possible by a human rater. A pre-prepared annotation template with controlled vocabularies in drop-down menus was used to reduce the variability in the annotations created [71]. The parameters annotated included: turning angle estimation (90°–360° in 45° increments, shown in Fig. 2) and duration of turn (seconds: milliseconds). Our definition of a turning episode is characterised by the initiation of pelvis rotation, continuing until the completion of the movement, which differs from a turn made within a walking arc, like walking around a table. The duration of labelled data recorded by the cameras for PD and C is 72.84 and 75.31 h, respectively.

Two clinicians annotated 50% of the turns each. Around 50% of the total number of annotations were cross-checked (randomly selecting 6 pairs from 12) by both clinician annotators, blinding the cross-checking clinician to the turning annotations produced by the other. Cohen's Kappa [72] statistic was calculated to evaluate inter-rater reliability. Any discrepancies were recorded, discussed, and resolved by the clinician raters, with a final review by a movement disorders specialist. The two clinician raters had an almost perfect [73] inter-rater agreement for turning angle annotations (Cohen's kappa = 0.96).

In addition to free-living movements, the turning clips in Turn-REMAP also include videos where the participants take part in clinical assessments and loosely-scripted activities (see Table 2). In the clinical assessments, participants underwent a series of predefined motor tasks that included completing walking and turning courses that are integral to the MDS-UPDRS (III) motor subscore [4]. Additionally, they were required to perform the timed-up-and-go (TUG) test [74] twice. Another task involved a 10-metre walk that incorporated three 180° turns, which participants carried out at their normal, fast, and slow paces. Naturally, the turning clips for these predefined 180° turns are labelled as 180°. Compared to free-living activities, the loosely-scripted activities consisted of food preparation tasks undertaken with only broad instructions and no one observing the participants.

**Table 2**
The number of turns, including their angles, in each type of activity in Turn-REMAP.

| Scenario | 90° | 135° | 180° | 225° | Total |
|---|---|---|---|---|---|
| Loosely-Scripted | 316 | 36 | 32 | 2 | 386 |
| Clinical Assessment | 7 | 1 | 41 | 0 | 49 |
| Free-living | 580 | 179 | 188 | 4 | 951 |
| Total | 903 | 216 | 261 | 6 | 1386 |

**Table 3**
Details for different bins on curated turning videos dataset Turn-H3.6M.

| Bin | # Videos | Min angle | Max angle | Avg. Angle | Avg. Duration |
|---|---|---|---|---|---|
| 45° | 372 | 45.2° | 67.5° | 55.3° | 1.1 s |
| 90° | 146 | 67.7° | 112.0° | 89.1° | 1.5 s |
| 135° | 59 | 115.0° | 156.4° | 134.9° | 2.4 s |
| 180° | 36 | 163.8° | 199.6° | 178.5° | 2.9 s |
| 225° | 6 | 213.2° | 234.7° | 222.2° | 3.2 s |
| **Total** | 619 | 45.2° | 234.7° | 79.6° | 1.5 s |

**Turn-H3.6M —** To further validate our proposed approach, we curated Turn-H3.6M, a specific turning action video subset of the Human3.6M benchmark [28] which consists of 3.6 million frames of RGB and 3D data of 11 professional actors performing various activities in a customised lab environment, such as walking a dog, smoking, taking a photo or talking on the phone. The dataset includes 3D human pose groundtruth data.

Previously, IMU-based turning estimation [34] has shown that compared to head, neck and ankle, sensory information on the lower back provides a more accurate estimation of turning angle. Following this, we used 3D groundtruth to locate frame sequences in the Human3.6M dataset with a consecutive hip rotation equal or larger to 45° (see example in Fig. 3). The 45° quantity corresponds to the increment between the angle labels within our bins and represents the minimum rotation required to classify a motion as a turning motion. The orientation of groundtruth hip joints serves as the groundtruth turning angle, and further enables the calculation of actual turning speeds, allowing for comparison with speeds derived from predicted angles.

We manually searched through the entire Human3.6M dataset and extracted 619 legitimate turning video clips at 50 fps, comprising a total of 45 199 frames. The clips have an average duration of 1.5 s, and the turning angle ranges from 45.2° to 234.7° (see Table 3).

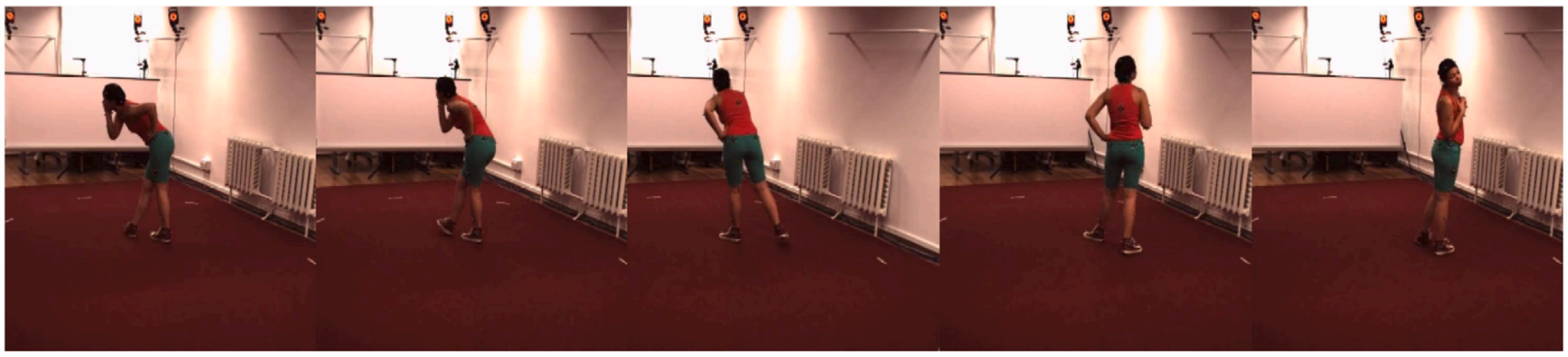

**Fig. 3.** A turning clip in Turn-H3.6M — An example extracted from the Human3.6M dataset [28].

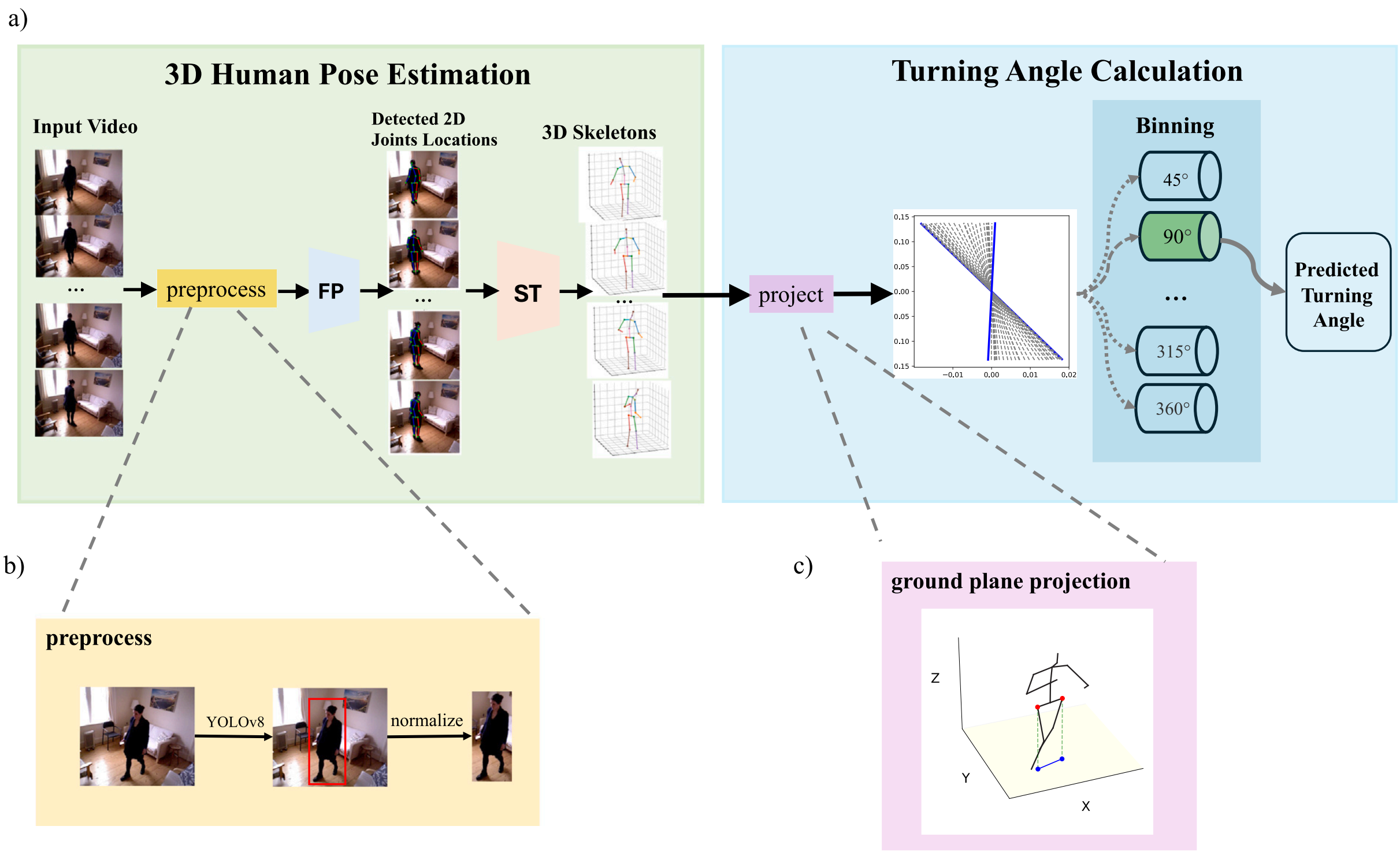

**Fig. 4.** The workflow of our proposed pipeline — (a) Using monocular videos as input, we apply FastPose (FP) [75] to estimate joint locations and lift the 2D skeleton series into 3D using the Strided Transformer (ST) [76]. With 3D skeletons, we compute the turning angle as the rotation of joints after being projected onto the horizontal plane. A continuous value of the angle is determined and further quantised into the nearest 45° bin, which is then compared to the clinician's annotation. (b) The preprocessing steps of video frames before feeding into FP. (c) The projection of the hip joints onto the ground plane as hip vector $\mathcal{H}_t$.

## 4. Methodology

In this section, we provide a detailed description of our proposed framework. Our overall pipeline has two major processes (Fig. 4): 3D human joints estimation and turning angle calculation.

**3D human joints estimation —** Our approach comprises a two-stage framework where we first detect 2D human joint locations in each frame of the video sequence and then reconstruct them in 3D space based on the spatial–temporal knowledge extracted from the temporal 2D skeletons series using a deep learning model. Another way of estimating 3D human pose from videos is to use a single deep learning model to infer the 3D coordinates from the RGB pixels directly in an end-to-end manner [77,78]. However, a more loosely coupled pipeline is chosen over end-to-end frameworks as it has been shown to achieve higher accuracy with gre lower computational cost on almost all of the benchmarks for human pose estimation [76,79,80].

To detect the 2D body joints in each video frame, we apply FastPose [75] as the 2D keypoints detector. The keypoints detector maps input video frames $\mathbf{V} \in \mathbb{R}^{T*W*H*3}$, into frames of 2D keypoint coordinates $\mathbf{K} \in \mathbb{R}^{T*J*2}$. $T$ is the number of frames of the video, $W$ and $H$ are the width and height of each frame and $J = 17$ is the number of joints (keypoints) in our skeleton, following the skeleton model from Human3.6M [28].

FastPose uses a top-down framework, which detects the human object from the frames and estimates the joint coordinates in the form of a heatmap within a bounding box. The model utilises the classical ResNet [58] as the image feature extraction backbone and then uses upsampling modules [81] and 1D convolution to generate heatmaps to represent the probability of each pixel being a human joint. FastPose outputs a heatmap for each joint, selecting the pixel with the maximum value as the joint's coordinate. FastPose does not directly regress human joint positions from an image; instead, it must be used in conjunction with a human object detector. In our experiment, we employed YOLOv8 as an off-the-shelf human object detector. The bounding boxes detected by YOLOv8 were used to crop the input image, and these cropped regions were then fed into FastPose for joints estimation. Before feeding the frames into FastPose, we apply standard preprocessing techniques [75,82]: rescaling, normalisation, and flip augmentation. The detected human bounding boxes are first rescaled to a uniform size of 256 × 196 resolution, as required by the model. Subsequently, the input is normalised by subtracting the mean pixel values for each RGB channel, which helps account for differences in brightness and contrast between frames. Pixel values are divided by 255 to normalise them to the range between 0 and 1. Additionally, we employ standard flip augmentation for both training and inference. In this process, we flip the input of FastPose to obtain a flipped output. By flipping the output back and averaging it with the original output, we derive the final prediction.

Having obtained 2D coordinates of human joints, we reconstruct the missing depth information to lift skeletons from 2D to 3D. This is inherently an ill-posed problem, as a single 2D skeleton could have been projected by an infinite number of different 3D poses. However, adding temporal knowledge on how the 2D skeleton changes over time could potentially lead to a more accurate 3D reconstruction.

Numerous architectures have been suggested to address this ill-posed problem [79,80]. We adopt the advanced model, Strided Transformer [76], to map the 2D keypoints series $\mathbf{K} \in \mathbb{R}^{T*J*2}$ into 3D skeleton $\mathbf{S} \in \mathbb{R}$. The Strided Transformer is a transformer-based architecture that converts 2D keypoints into 3D groundtruth using the original transformer encoder [83]. The output is then processed by another transformer encoder with strided convolutions, aggregating the sequence to reconstruct the 3D joints of the centre frame. Notably, the Strided Transformer introduces extra constraints to ensure temporal smoothness while simultaneously aggregating long-range information across the skeleton sequence.

Partial occlusions that are not severe are handled by the 2D keypoint detector FastPose by generating a plausible prediction of missing joint locations. As a result, a complete 2D skeleton is provided as a legitimate input to the Strided Transformer, which then reconstructs the 3D skeleton. Additionally, the Strided Transformer uses the context from surrounding frames to predict 3D joint locations in a central frame of a 27-frame sequence. When a partial occlusion occurs, this temporal smoothness constraint prevents drastic pose changes and helps estimate the joint's 3D position using information from adjacent frames.

In summary, given an RGB video, the pipeline detects the location of joints on each frame and projects a time series of 2D human skeletons as input into a reconstruction model trained with 3D motion capture groundtruth. The final output is then a time series of 3D human skeletons for each turning video clip.

**Turning angle estimation —** The availability of 3D coordinates for skeleton joints, spanning from the head to the feet, offers the flexibility to conduct precise quantitative assessments of various movements. However, in the context of turning analysis, it is important to note that while the concept of turning has been previously defined, the specific definition of its magnitude or angle has not been explored in prior research focused on skeleton-based gait analysis. In our methodology, the frontal plane is selected over the sagittal or transverse planes to calculate the angle of turning, for it is the anatomical landmark of the human body. The hip and knee joints on the frontal plane, when the human body is in an upright position, are used to estimate the turning angle in a plane parallel to the assumed flat, ground plane, denoted as the $XY$ plane.

The hip and knee vectors $\mathcal{H}_t$, $\mathcal{K}_t$ respectively, at frame $t$ are defined as

$$\mathcal{H}_t = (x_t, y_t)_{left_hip} - (x_t, y_t)_{right_hip} \ , \quad (1)$$

$$\mathcal{K}_t = (x_t, y_t)_{left_knee} - (x_t, y_t)_{right_knee} \ . \quad (2)$$

For a turning video with $T$ frames, we calculate the angle $\theta$ between the corresponding vectors of two consecutive frames $t$ and $t+1$ for the knee and hip joints, and then sum and average the two angles as

$$\theta = \frac{1}{2}\sum_{t=0}^{T-2}\left(\sin^{-1}\left(\frac{\|\mathcal{H}_t \times \mathcal{H}_{t+1}\|}{\|\mathcal{H}_t\| \ \|\mathcal{H}_{t+1}\|}\right) + \sin^{-1}\left(\frac{\|\mathcal{K}_t \times \mathcal{K}_{t+1}\|}{\|\mathcal{K}_t\| \ \|\mathcal{K}_{t+1}\|}\right)\right) \ . \quad (3)$$

For our trimmed videos with duration $d$, the angular speed $\omega$ is subsequently computed as

$$\omega = \theta / d \ . \quad (4)$$

In ablations, we consider the shoulder, hip and knee joints, together and separately, and show that the combination of hip and knee vectors performs best. The proposed turning angle estimation algorithm acts as a plug-in-and-play component for any 3D pose estimation model that produces 3D skeletons, providing a compatible method for future comparison. For each video clip, in terms of calculating the overall angle, it is mathematically the same as using only the first and last frame vectors, but the frame-by-frame manner could also inform us of how the velocity changes within one turning motion. To evaluate the performance of turning angle estimation, we select two categorical metrics (accuracy, weighted precision) and one regression metric (mean absolute error) against the clinician's annotation. These metrics enable us to assess both the performance of the discrete classification and quantify the magnitude of deviation between estimated and annotated turning angles.

## 5. Experiments

**Implementation and Evaluation —** Experiments are conducted in PyTorch on one single NVIDIA 4060Ti GPU and a 12-core AMD Ryzen 5 5500 CPU. We evaluate our proposed method via three key metrics: accuracy, Mean Absolute Error (MAE) in degrees, and weighted precision (WPrec). Accuracy assesses the proportion of predicted angles that correctly fall into their respective bins, showing the categorical correctness of our predictions. MAE is calculated as the average of the absolute difference between the predicted values for angles, as well as speed in Turn-H3.6M, against groundtruth. WPrec measures the percentage of true positive predictions for all positive predictions across angle bins, weighted by each bin's sample size [84]. For example, if a turn is predicted as 90°, WPrec indicates the probability that the actual turn is 90°.

**Pretraining Datasets —** In our pipeline, the utilised FastPose model is trained on the MSCOCO [85] pose estimation dataset, which provides multi-person keypoint annotations with 150K person instances in 57K images of complex everyday scenes containing common objects in their natural context. The Strided Transformer is trained on Human3.6M [28], consisting of 3.6 million images featuring 7 healthy actors performing 15 everyday activities, as well as the 2D and 3D ground truth human joint locations using an accurate motion capture system in a controlled environment. The Strided Transformer is trained on subjects S1, S5, S6, S7 and S8 of Human3.6M, while subject S9 and S11 are held out for validation following the standard setup of related literature [79,80]. Due to the constraints of an unobtrusive, free-living data collection environment, we did not perform any fine-tuning on Turn-REMAP. Instead, we investigated the generalisation performance of the angular error across three data settings: (1) unseen test data from the challenging, real-world Turn-REMAP; (2) seen training data from subjects S1, S5, S6, S7 and S8 in Turn-H3.6M; and (3) unseen validation data from subjects S9 and S11 in Turn-H3.6M.

**Results on Turn-REMAP —** We compared the predicted turning angle against the clinician's annotations for Turn-REMAP. Based on the rotation of hip and knee joints, our method correctly estimates the angle for 41.6% of all the turns on average, with an overall MAE of 34.7° and WPrec of 68.3% across 1386 videos.

We investigated turning in Turn-REMAP by the turning scenario, location of the turn and subject's condition. Table 4a reports the accuracy under the three scenarios of loosely scripted, clinical, and free-living. Our model across these scenarios yields an accuracy ranging from 26.5% to 44.0%, an MAE ranging from 33.4° to 59.2° and WPrec ranging from 66.2% to 79.1%, with overall averages of 36.0%, 42.5° and 71.7%, respectively. The performance on turns that happened during clinical assessments is significantly worse than the other two scenarios, marking it an outlier. This is largely due to the heightened occurrence of self-occlusion, which hampers the quality of the reconstructed skeleton. Notably, 40 out of 49 turns under clinical assessment

**Table 4**
%Accuracy$_\theta$, MAE$_\theta$ (°) and %WPrec$_\theta$ of angle estimator against groundtruth on Turn-REMAP — There are a total of 1386 turns in Turn-REMAP. Accuracy is calculated by quantising the predicted angle into the nearest 45° bin. MAE$_\theta$ shows the average error between the predicted continuous values and our labels. WPrec is the weighted average of each bin, with weights based on the sample size.

(a) Grouped by types of activities. Scripted stands for loosely scripted activities; Clinical stands for clinical assessment.

| Metrics | Scripted | Clinical | Free-living | Avg. |
|---|---|---|---|---|
| Accuracy$_\theta$ | 37.6 | 26.5 | 44.0 | 36.0 |
| MAE$_\theta$ | 34.8 | 59.2 | 33.4 | 42.5 |
| WPrec$_\theta$ | 79.1 | 79.7 | 66.2 | 71.7 |
| # Turns | 386 | 49 | 951 | 462 |

(b) Grouped by location.

| Metrics | Din. | Hall | Kit. | Liv. | Stairs | Avg. |
|---|---|---|---|---|---|---|
| Accuracy$_\theta$ | 35.9 | 36.0 | 42.9 | 38.4 | 40.0 | 38.6 |
| MAE$_\theta$ | 41.3 | 38.4 | 33.0 | 35.2 | 21.7 | 33.9 |
| WPrec$_\theta$ | 71.4 | 75.2 | 70.3 | 59.9 | 80.0 | 71.4 |
| # Turns | 92 | 89 | 1062 | 138 | 5 | 277 |

(c) Grouped by PD and C.

| Metrics | PD | C | Avg. |
|---|---|---|---|
| Accuracy$_\theta$ | 42.0 | 41.0 | 41.5 |
| MAE$_\theta$ | 34.4 | 35.1 | 34.8 |
| WPrec$_\theta$ | 68.2 | 68.7 | 68.5 |
| # Turns | 747 | 639 | 693 |

are participants performing the predefined 180° turns of the TUG test in the narrow hallway.

Table 4b shows that the performance of our model for turns across different locations remains fairly consistent, with the accuracy ranging from 35.9% to 42.9% and an average accuracy of 38.6%. There is a wide range of variation for MAE spanning from 21.7° to 41.3° and an average of 33.9° and a contributing factor to these results is how certain spaces are defined within Turn-REMAP. The kitchen, living room, and stairs are captured as open spaces with no occlusion from furniture, resulting in lower MAE and higher accuracy for turns in these areas. In contrast, the dining room and hallway show increased MAE and reduced accuracy due to frequent occlusions from a centrally located table in the dining room and self-occlusion in the hallway. WPrec ranges from 59.9% to 80.0%, with an average of 71.4%. Finally in Table 4c, we observe only marginal difference between subjects with PD, who had an accuracy of 42.0%, an MAE of 34.4° and WPrec of 68.2%, and control subjects, who had an accuracy of 41.0%, MAE of 35.1° and WPrec of 68.7%.

**Results on Turn-H3.6M —** The availability of 3D groundtruth in our curated dataset allows us to calculate the actual turning angle and turning speed, facilitating a direct comparison against the predictions of our model. Our proposed approach on the entire Turn-H3.6M dataset yields an average accuracy of 73.5% and an MAE of 18.5° for angle prediction, with a turning speed MAE of 15.5°/s and a WPrec of 86.2%.

As shown in Table 5a the proposed method yields an accuracy ranging from 50.0% to 80.6% and an MAE ranging from 13.4° to 20.7° with averages of 71.6% and 16.1°, respectively, across different turning angle bins. The MAE for turning speed ranges from 5.3 °/s to 16.9 °/s and improves for larger turning angles possibly because larger turns may exhibit more pronounced changes in speed. We investigate the performance of our pipeline on different subjects in Table 5b. Following previous works, such as [79,80], our model is trained on subjects S1, S5, S6, S7, and S8, while S9 and S11 are held for testing. For turning angle prediction, the accuracy spans from 63.2% to 80.0%, while the MAE varies between 13.3° and 24.7°, with respective averages across different subjects being 74.1% and 17.8°. The MAE for turning speed spans from 6.9°/s to 25.3°/s, with an average of 13.8°/s across different subjects. The WPrec ranges from 75.2% to 93.9% with an average of 85.9%. Although not included in the training phase, the performance of our model on test subjects S9 and S11, in terms of turning angle and speed calculation, falls within the consistent range observed for the other subjects used in training, suggesting the potential for generalisation to previously unseen data. The performance of the turns in S7 stands out as an outlier, showing the poorest results for both speed and angle. A possible explanation could be that the turns of S7 have the lowest average turning angle compared to those of all other subjects. Specifically, 113 out of 144 turns are at 45°, an angle at which our model tends to underperform (Table 5a).

In Table 5c, we see the results of turning angle prediction for turns while the subject performs different actions. Accuracy fluctuates between 63.3% and 84.8%, and MAE spans a range of 12.5° to 26.1°, yielding average values of 75.1% for accuracy and 17.7° for MAE. Our predicted turning speed shows an MAE ranging from 9.3°/s to 21.0°/s, with an average of 14.5°/s. WPrec ranges from 86.2% to 96.1% with an average of 87.9%. The original purpose of these pre-defined activities is to elicit various and diverse human body poses. Although there are imbalanced numbers of turns in different activities, the difference in performance can be attributed to the dynamics of movement including speed and motion pattern.

**Comparison between performance on Turn-REMAP and Turn-H3.6M —** We further illustrate a comparison of the distribution of turns across different angle labels in the Turn-REMAP dataset against the distribution of turns at the same angles in the Turn-H3.6M dataset in Fig. 5(a). The values in Turn-H3.6M are closer to the expected bin angles, while in Turn-REMAP, the predicted angles tend to be underestimated. In these examined bins labelled as 90°, 135°, 180° and 225° in Fig. 5(a), the standard deviations for the Turn-REMAP dataset are 35.3°, 37.6°, 46.4° and 70.0°, compared to those for the Turn-H3.6M dataset at 22.2°, 20.5°, 16.8° and 21.5°, respectively. This shows that, compared to Turn-REMAP, there is less variability and uncertainty within each bin for predictions in Turn-H3.6M. The difference in performance is further shown in Fig. 5(b), where we find that the distribution of MAE for the Turn-REMAP has a wider spread, while in Turn-H3.6M, 89.5% of the errors are smaller than 40°.

**Cross-sectional analysis of free-living turning characteristics —** The turning angle serves as a fundamental metric for calculating more complex turning characteristics, such as maximum angular velocity. Here, we demonstrate the potential of the proposed method via a use case of PD classification from healthy controls. Since the turning angle is computed on a frame-by-frame basis, the maximum angular velocity $w_{max}$ of a turn can be calculated as

$$w_{max} = \max(\theta_0, \theta_1, \dots, \theta_t, \dots, \theta_{N-2}) \times dt \;, \tag{5}$$

where $dt$ represents the time interval for a single frame, calculated as the reciprocal of the frames per second (fps), and $\theta_t$ is the angle between two consecutive frames $t$ and $t+1$, where $t \in [0, N-2]$, and $N$ represents the number of frames, i.e.

$$\theta_t = \frac{1}{2}\left(\sin^{-1}\left(\frac{\|\mathcal{H}_t \times \mathcal{H}_{t+1}\|}{\|\mathcal{H}_t\|\ \|\mathcal{H}_{t+1}\|}\right) + \sin^{-1}\left(\frac{\|\mathcal{K}_t \times \mathcal{K}_{t+1}\|}{\|\mathcal{K}_t\|\ \|\mathcal{K}_{t+1}\|}\right)\right) . \tag{6}$$

In the Turn-REMAP cohort, we compared individuals with and without PD (age-controlled [18]). By applying our model pretrained on Human3.6M to this cohort, we conducted a cross-sectional statistical analysis of the estimated turning angles and max angular velocities (with results summarised in Table 6). We consider turns from different subjects to be independent, and the mean values reported below are averaged across subjects within each group. The mean turning angle

**Table 5**
%Accuracy$_\theta$, MAE$_\theta$ (°), MAE$_\omega$ (°/s), and %WPrec$_\theta$ on Turn-H3.6M — There are a total of 619 turns in Turn-H3.6M. Avg. represents the mean value across these groups. Avg$_\omega$ and Avg$_d$ represent the average turning speed and average duration within these groups.

(a) Grouped by turning angle.

| | 45° | 90° | 135° | 180° | 225° | Avg. |
|---|---|---|---|---|---|---|
| Accuracy$_\theta$ | 70.2 | 79.5 | 78.0 | 80.6 | 50.0 | 71.6 |
| MAE$_\theta$ | 20.7 | 15.4 | 15.3 | 13.4 | 15.8 | 16.1 |
| MAE$_\omega$ | 19.6 | 11.3 | 7.0 | 5.3 | 5.4 | 9.7 |
| # Turns | 372 | 146 | 59 | 36 | 6 | 124 |

(b) Grouped by subjects.

| | Train | | | | | Validation | | Avg. |
|---|---|---|---|---|---|---|---|---|
| | S1 | S5 | S6 | S7 | S8 | S9 | S11 | |
| Accuracy$_\theta$ | 64.1 | 75.9 | 80.0 | 63.2 | 78.6 | 76.7 | 80.0 | 74.1 |
| MAE$_\theta$ | 21.6 | 16.2 | 16.9 | 24.7 | 13.3 | 17.3 | 14.8 | 17.8 |
| MAE$_\omega$ | 15.8 | 12.3 | 14.3 | 25.3 | 6.9 | 14.3 | 7.8 | 13.8 |
| WPrec$_\theta$ | 75.2 | 84.6 | 93.9 | 93.7 | 84.8 | 84.9 | 84.3 | 85.9 |
| # Turns | 39 | 116 | 95 | 144 | 56 | 129 | 40 | 88 |

(c) Grouped by types of actions — from left to right: giving directions, eating, greeting, talking on the phone, posing, discussing, smoking, walking, waiting and taking photos.

| | Direc. | Eat. | Greet. | Phon. | Pos. | Disc. | Smok. | Walk. | Wait. | Photo | Avg. |
|---|---|---|---|---|---|---|---|---|---|---|---|
| Accuracy$_\theta$ | 76.2 | 63.3 | 72.7 | 74.6 | 82.4 | 73.9 | 72.4 | 72.9 | 84.8 | 77.3 | 75.1 |
| MAE$_\theta$ | 15.9 | 26.1 | 15.6 | 19.1 | 19.8 | 15.8 | 18.4 | 19.0 | 12.5 | 14.4 | 17.7 |
| MAE$_\omega$ | 13.0 | 21.0 | 12.6 | 14.9 | 18.0 | 12.2 | 16.2 | 16.6 | 9.3 | 11.1 | 14.5 |
| WPrec$_\theta$ | 86.2 | 90.0 | 87.0 | 86.3 | 96.1 | 79.6 | 91.6 | 86.2 | 90.3 | 85.5 | 87.9 |
| # Turns | 21 | 49 | 33 | 71 | 17 | 46 | 76 | 251 | 33 | 22 | 62 |

**Table 6**
Group differences in turning angle and maximum angular velocity between PD and control subjects.

| Measure | Group | Mean ± SD | $t$-statistic ($p$) | Effect Size (Cohen's $d$) |
|---|---|---|---|---|
| Turning angle (°) | PD<br>Control | 92.65 ± 13.21<br>103.75 ± 16.75 | −1.73 ($p$ = 0.0998) | −0.74 |
| Max angular velocity (°/s) | PD<br>Control | 127.86 ± 29.77<br>160.19 ± 36.49 | −2.28 ($p$ = 0.0339) | −0.97 |

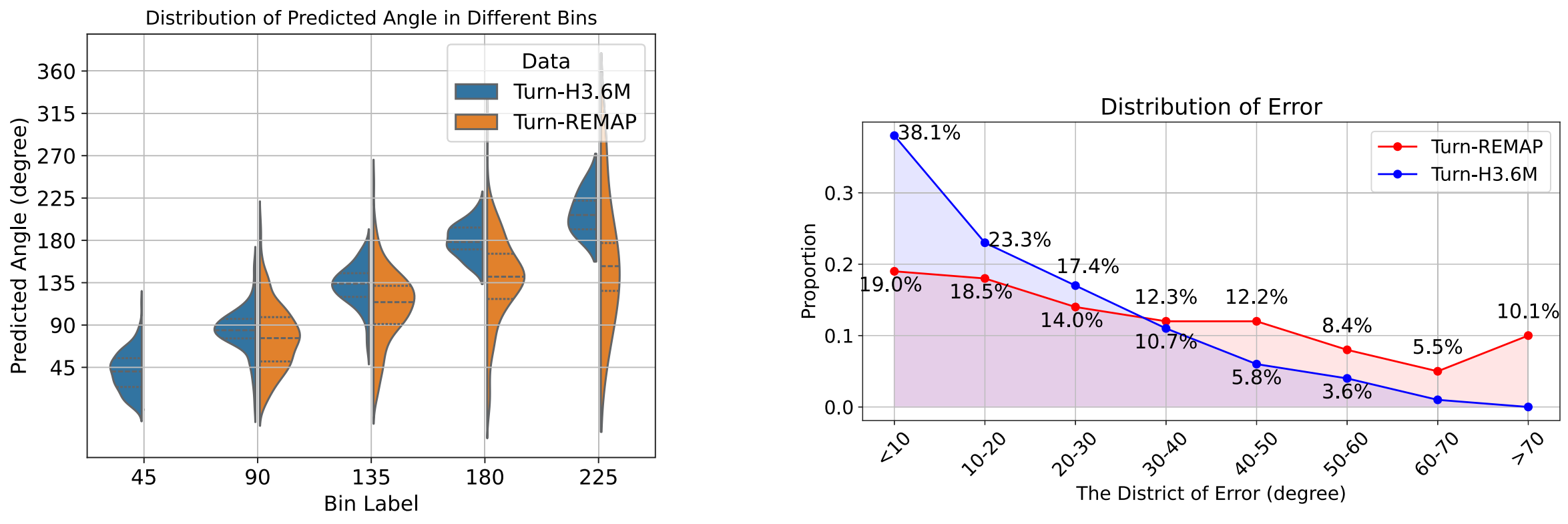

**(a)** Comparison of distribution of predicted angle    **(b)** Comparison of distribution of MAE

**Fig. 5.** Comparison of performance of the proposed model on Turn-REMAP and Turn-H3.6M — In (a), mean predicted angles for Turn-REMAP are 76.1°, 114.3°, 141.4°, and 152.6° at bins 90°, 135°, 180°, and 225°, respectively; for Turn-H3.6M, they are 84.4°, 133.9°, 178.2°, and 207.0°.

was 92.65° ± 13.21° for the PD group and 103.75° ± 16.75° for the control group. A two-tailed independent-samples $t$-test yielded a $t$-statistic of −1.73 ($p$ = 0.0998), with a mean difference of −11.10° (95% CI: [−24.56°, 2.36°]). Although this difference was not statistically significant, the effect size was moderate (Cohen's $d$ = −0.74), suggesting a potentially meaningful group difference. For maximum angular velocity, the average per-subject value was 127.86°/s ± 29.77°/s for the PD group and 160.19°/s ± 36.49°/s for the control group. In this case, a two-sample $t$-test revealed a statistically significant difference, with a $t$-statistic of −2.28 ($p$ = 0.0339), a mean difference of −32.33°/s (95% CI: [−62.03°/s, −2.63°/s]), and a large effect size (Cohen's $d$ = −0.97).

## 6. Ablations

The accurate detection of 2D skeleton keypoints in each frame of our input clips is an significant contributor to the overall accuracy of our method. Another fundamental concern is which single or combination of 'body parts' should be engaged for the computation of the turning angle. We investigate these two issues in our ablation study.

**The effect of different 2D keypoints —** We investigate how various 2D keypoint detectors impact the performance of the turning angle estimation in Turn-H3.6M. We applied SimplePose [86], HRNet [82] and FastPose [75] as prospective 2D keypoint detectors respectively and evaluated their performance in estimating turning angles. All three models were trained on the MSCOCO dataset [85] following the same settings. HRNet and FastPose were chosen because they are advanced 2D keypoint detection models, while SimplePose, with its minimal yet effective design, was chosen to determine if more complex models are only overfitting the training dataset.

**Table 7**
The effect of different 2D keypoints input on 3D reconstruction performance — %Accuracy$_\theta$ and MAE$_\theta$ (°) evaluate the performance of turning parameters estimation on Turn-H3.6M. Params shows the number of trainable parameters in the models (in millions), and GFLOPs (Giga Floating Point Operations Per Second) shows the computational cost of a single forward pass during inference on Turn-H3.6M.

| 2D Keypoints Input | Accuracy$_\theta$ | MAE$_\theta$ | Params | GFLOPs |
|---|---|---|---|---|
| With SimplePose | 71.6 | 18.5 | 34.0M | 406.9 |
| With HRNet | 71.4 | **18.4** | 63.6M | 674.0 |
| With FastPose | **73.5** | 18.5 | 40.5M | **246.7** |

**Table 8**
The effect of using different combinations of joints on Turn-REMAP — Using the combination of hip and knee joints yields the best overall %accuracy, using all three sets of joints yields the best overall MAE$_\theta$.

| Selected Joints | Accuracy$_\theta$ | MAE$_\theta$ |
|---|---|---|
| hip | 39.7 | 36.3 |
| knee | 36.7 | 37.4 |
| shoulder | 38.5 | 36.4 |
| hip+knee | **41.6** | 34.7 |
| hip+shoulder | 40.3 | 35.7 |
| knee+shoulder | 41.1 | 34.4 |
| hip+knee+shoulder | 41.5 | **34.3** |

The MAE of these three models does not vary significantly, with values at 18.4° and 18.5°. Among them, FastPose offers the highest accuracy and significantly reduces computational costs in detecting 2D keypoints (see Table 7).

**The effect of using different joints —** We also calculated the turning angle using different combinations of knee joints, shoulder joints and hip joints to determine which body part provides the best turning angle estimation. We chose to perform this ablation on Turn-REMAP instead of Turn-H3.6M because the groundtruth for turning angles in Turn-REMAP is derived from the clinical expertise of movement disorder specialists. In contrast, the joints used to determine the turning angle groundtruth in Turn-H3.6M have already been discussed and defined.

On the human frontal plane, similar to knee and hip joints, shoulder joints are also potentially good indicators of the orientation of the body [87]. However, PD patients have difficulty in maintaining lateral balance during weight shifts from one foot to the other while turning, demonstrating a greater inclination angle in the frontal plane than healthy controls [88]. This suggests that, in PD, the shoulder joints may become less reliable for initiating turns, whereas combining the hip and knee joints shows less variability and may remain more stable in an upright stance.

As shown in Table 8, the average predicted angle using both hip joints and knee joints yields the best accuracy, while averaging all three sets of joints gives the lowest MAE for turning angle.

## 7. Discussion

Previous methods for turning analysis have been developed primarily for laboratory or clinical settings to evaluate scripted activities [27,89–91]. In Turn-REMAP, we record gait videos in a home-like, unobtrusive environment with PD and control subjects, and provide quantitative evaluations on the accuracy and estimation errors of turning angles during free-living activities. Pham et al. [26] also measured turning in a free-living environment, however, their method measured turning angles from IMUs alone, while our method is video-based. Pham et al. [26] recorded videos to manually validate their estimated results and report an overall error of 0.06°, but we contend that estimating turning angles at an accurate enough resolution to achieve such low error measurements by examining videos with the naked eye is unreliable. Turning angle could also be calculated from 3D skeletons reconstructed using automatic methods [80,92], whose performance is evaluated using Mean Per Joint Position Error (MPJPE) and Mean Per Joint Angle Error (MPJAE). Naturally, the granularity and resolution of the estimated turning angle depend on these performance metrics as well as the video's frame rate. Though predicting the exact turning angle offers greater sensitivity, the increased precision often requires specialised setups or additional costs. In our free-living scenario, automatically binning turning angles into discrete values can save clinicians significant time and effort in PD assessments — for example, consistently comparing 180° turns can help reveal early signs of disease progression in Parkinson's disease [23]. Some other IMU-based studies [19,21,93] have also extended their methodology to home environments, but none of these studies validated the measurement accuracy in the free-living setting.

Although the overall measurement accuracy of our Turn-REMAP dataset is not yet robust enough for clinical diagnosis, it establishes a baseline for future, passive, video-based analysis of turning movements in indoor, free-living environments. Our manual examination of incorrectly classified video clips and their corresponding 3D skeletons revealed that depth reconstruction ambiguity is an important factor [79,80] that affects the accurate calculation of turning angles. Recovering the missing depth from a 2D image is inherently an ill-posed problem as infinitely many 3D poses can project to the same 2D skeleton. Despite our utilised models being pretrained on large-scale laboratory 3D motion data, generalising this performance to reconstruct unseen poses in our in-the-wild PD turning dataset remains a substantial challenge.

In Turn-REMAP, we find that the performance of our method on turns in free-living and loosely scripted activities is better than in clinical assessment (Table 4a). The reason for the performance degradation on these turns during the clinical assessment is the heightened occurrence of self-occlusion, where 40 of 49 of these turns are scripted 180° turns in a narrow hallway. This is confirmed by the findings in Table 4b, which shows that locations like the dining room and hall, which have more occlusions, tend to have lower accuracy and higher MAE. Additionally, we find there is no large difference in the performance of predicting turning angles for PD and Control (Table 4c), suggesting special PD gait characteristics do not affect the performance of our method. In contrast, IMU-based turning measurement methods [19,21,23,36] rely heavily on setting thresholds of angular velocity and relative orientation of the sensor attached to a single body part. Compared to skeleton-based models, the isolated sensory kinematic parameters are more easily affected by common PD symptoms, such as freezing of gait and slow turning speed [23]. However, a disadvantage of video-based methods is that, aside from occlusions — such as when the subject's entire body is not in the line of sight — free-living scenarios may also introduce other challenges, including multi-person interactions. These could potentially be addressed using technology such as person re-identification for at-home health monitoring [94,95].

The comparative statistics of the performance on Turn-REAMP and Turn-H3.6M reveal challenges in generalising our pretrained human pose estimation model from the lab-based Turn-H3.6M dataset to the diverse, in-the-wild Turn-REMAP dataset. Different global position distributions [96], camera parameters [97], and diverse human body sizes and shapes, as well as articulated movements [98,99] highlight the need to enhance model robustness and adaptability to better handle real-world variability. To bridge this gap and enhance the ability to generalise to new, unseen data, it is crucial to implement domain adaptation strategies in deep learning models and conduct cross-dataset validation. Although there is still room for improvement in the accuracy of turning angle estimation, our cross-sectional analysis reveals a statistically significant group difference between PD and C in maximum angular velocity. Specifically, PD subjects exhibited much slower turns and subtle reductions in turning angles, which is consistent with previous studies in different PD cohorts [19,44]. This demonstrates that, even though our model was pretrained on public benchmark datasets composed of subjects without Parkinson's disease, it was still able to characterise the impaired gait of older, early-stage PD subjects in Turn-REMAP.

Our model performs at 73.5% (Table 7) accuracy on Turn-H3.6M. This performance is limited by the inherent design of existing pose estimation algorithms, which are not specifically engineered to tackle biomechanical challenges, such as the analysis of turning characteristics. The training of these 2D–3D lifting models is usually guided by Mean Per Joint Position Error (MPJPE) [28] loss, which focuses on minimising the absolute distance between the locations of the groundtruth joints and the predictions. However, this criterion does not sufficiently address the requirements for temporal smoothness or accurate angular estimation. Therefore, further work on turning analysis involves building a downstream turning analysis algorithm based on the extracted deep learning features.

## 8. Conclusion and future work

Continuously and automatically measuring turning characteristics in a free-living environment could enhance the current clinical rating scale by capturing the true motor symptoms which fluctuate hour by hour. This study is the first effort to detect the fine-grained angle of turn in gait using video data where people are unscripted and in a home setting. In this paper, we introduced the Turn-REMAP and Turn-H3.6M datasets. Turn-REMAP is the first dataset of free-living turning movements that includes clinician-annotated, quantised turning angle groundtruth for both PD patients and control subjects across various scenarios and locations. Turn-H3.6M is derived from the lab-based, large-scale 3D pose benchmark known as Human3.6M, curated specifically for turning data analysis. To estimate the turning angle of a subject in raw RGB videos, we utilised a deep learning framework to reconstruct human joints in 3D space. We then proposed a turning angle calculation approach based on joint rotation. Our framework was applied to the unique Turn-REMAP dataset and further validated on Turn-H3.6M.

While the accuracy of our models may not yet allow their application in the real world, they nevertheless establish a previously non-existent baseline and offer valuable insights for future video-based research in challenging free-living scenarios. Our sample size of 24 people, including 12 people with PD, demonstrates that our approach to detecting turning angles is promising and provides a proof of concept. Automatically computing turning angles in a free-living environment is foundational for future longitudinal, in-home monitoring of PD. There are many potential avenues to build upon our work for more accurate turning angle estimation. Although Turn-REMAP and Turn-H3.6M only consist of trimmed turning clips, our methods can be extended to untrimmed videos. We could also infer additional turning metrics such as turning speed from the estimated turning angle. These metrics can be used to classify PD and control subjects, infer clinical rating scores of disease severity, or assess on/off medication status in free-living video recordings. Another extension for more accurate turning angle computation could be to replace our skeleton model with other models, such as via Human Mesh Recovery [100] which could offer additional parameters for turning angle estimation.

## CRediT authorship contribution statement

**Qiushuo Cheng:** Writing – review & editing, Writing – original draft, Visualization, Validation, Methodology, Investigation, Formal analysis, Data curation, Conceptualization. **Catherine Morgan:** Writing – review & editing, Writing – original draft, Supervision, Resources, Methodology, Formal analysis, Data curation. **Arindam Sikdar:** Methodology, Data curation, Conceptualization. **Alessandro Masullo:** Supervision. **Alan Whone:** Supervision, Project administration. **Majid Mirmehdi:** Writing – review & editing, Supervision, Project administration, Methodology.

## Declaration of competing interest

The authors have no conflict of interest in this work.

## Acknowledgements

The authors gratefully thank the study participants for their time and effort in participating in this research. We also acknowledge the local Parkinson's and other Movement Disorders Health Integration Team (Patient and Public Involvement Group) for their assistance at each step of the study design. This work was supported by the SPHERE Next Steps Project funded by the EPSRC (grant EP/R005273/1), the Elizabeth Blackwell Institute for Health Research at the University of Bristol, the Wellcome Trust Institutional Strategic Support Fund (grant 204813/Z/16/Z), Cure Parkinson's Trust, United Kingdom (grant AW021), and by IXICO (grant R101507-101). Dr Jonathan de Pass and Mrs Georgina de Pass also made a charitable donation to the University of Bristol through the Development and Alumni Relations Office to support research into Parkinson's Disease. The first author was supported by the Engineering and Physical Sciences Research Council Digital Health and Care Centre for Doctoral Training at the University of Bristol (UKRI Grant No. EP/S023704/1).